# An Adaptive Population Size Differential Evolution with Novel Mutation Strategy for Constrained Optimization

Yuan Fu, Hu Wang, and Meng-Zhu Yang

*Abstract*—Differential evolution (DE) has competitive performance on constrained optimization problems (COPs), which targets at searching for global optimal solution without violating the constraints. Generally, researchers pay more attention on avoiding violating the constraints than better objective function value. To achieve the aim of searching the feasible solutions accurately, an adaptive population size method and an adaptive mutation strategy are proposed in the paper. The adaptive population method is similar to a state switch which controls the exploring state and exploiting state according to the situation of feasible solution search. The novel mutation strategy is designed to enhance the effect of status switch based on adaptive population size, which is useful to reduce the constraint violations. Moreover, a mechanism based on multipopulation competition and a more precise method of constraint control are adopted in the proposed algorithm. The proposed differential evolution algorithm, APDE-NS, is evaluated on the benchmark problems from CEC2017 constrained real parameter optimization. The experimental results show the effectiveness of the proposed method is competitive compared to other state-of-the-art algorithms.

## I. INTRODUCTION

DIFFERENTIAL evolutions (DE) attracts lots of researchers attention because of its competitive performance on constrained optimization problems over the past decades [1]-[4]. Moreover, many optimization problems in the real world are regarded as constrained optimization problems (COPs), which may lead to more strict convergence conditions. The core of studying constrained optimization lies in how to search for feasible solutions accurately. Since DE is proposed by Storn and Price [6], many efforts have been designed to enhance the effect of DE on constrained problem, which mainly focused on the setting of control parameters such as scaling factor ($F$), crossover rate ($Cr$), selection mechanism, mutation strategies, and population size ($Np$).

Many experiments and analysis have proved DE is a parameter-dependence algorithm, therefore, the most suitable parameter setting may vary when solving different problems, however, the specific interaction between performance and parameter is still intricate. It also means that tuning the parameter manually might need expensive computation. This consideration motivated researchers to design the adaptive parameter setting method such as SaDE [7], JADE [8], jDE [9], SHADE [10], IDE [40], and JADE_sort [11]. The core of the above adaptive parameter control techniques is tuning the current generation parameters according to feedbacks collected in the previous evolutionary process. The targets of adaptive controlling are $Cr$ and $F$ in most studies, the quantity of study on the adaptive population size is relatively small. In the literatures, another parameter setting technique is also widely used such as EPSDE [12], Adaptive DE [13], CoDE [14], and UDE [36] which select the parameters from a pool contains sets of discrete candidate values. The methods mentioned above have been proved to significantly improve the performance of the algorithm.

Selection mechanism is also important to the effect of algorithm, which is a mechanism based on greedy choice in the traditional DE. This greedy feature accelerates the convergence speed. Whether the target or the trail (offspring) survives during the evolution process is determined by the selection mechanism, which is described as

$$x_i^{(t+1)} = \begin{cases} u_i^{(t)} & if\ f(u^{(t)}) \leq f(x_i^{(t)}) \\ x_i^{(t)} & otherwise \end{cases} \quad (1)$$

where $f(.)$ is the objective function to be optimization. $u^{(t)}$ is the $i^{th}$ individual trail vector in $t^{th}$ generation. $x_i^{(t)}$ is the $i^{th}$ individual target vector, The "≤" is helpful to guide the DE population to a fitness landscape and reduce the probability of evolutionary process becoming stagnated [2]. In constrained optimization problems, both the constraint violations and the objective function value should be considered, however, the classical selection mechanism only focuses on the objective function value. Many researchers have modified the classical mechanism in order to enhance algorithm adaptability for the special requirements of constrained optimization solution properties, such as superiority of feasible solutions (SF) [15, 16].

Manuscript received … This work has been supported by Project of the Program of National Natural Science Foundation of China under the Grant Numbers 11572120, and the National Key Research and Development Program of China under the Grant Number 2017YFB0203701. *(Corresponding author: Hu Wang)*

Y. Fu, H. Wang, and M.-Z. Yang are with the State Key Laboratory of Advanced Design and Manufacturing for Vehicle Body, School of Mechanical and Vehicle Engineering, Hunan University, Changsha 410082, China. (E-mail: wanghu@hnu.edu.cn; fuy@hnu.edu.cn)

Mutation strategies directly affect the balance of exploration and exploitation during the evolutionary process, some most frequently referred mutation strategies are listed below:

"DE/rand/1":
$$v_i^{(t)} = x_{R_1^i}^{(t)} + F\left(x_{R_2^i}^{(t)} - x_{R_3^i}^{(t)}\right) \quad (2)$$
"DE/best/1":
$$v_i^{(t)} = x_{best}^{(t)} + F\left(x_{R_1^i}^{(t)} - x_{R_2^i}^{(t)}\right) \quad (3)$$
"DE/current-to-best/1":
$$v_i^{(t)} = x_i^{(t)} + F\left(x_{best}^{(t)} - x_i^{(t)}\right) + F\left(x_{R_2^i}^{(t)} - x_{R_2^i}^{(t)}\right) \quad (4)$$

Where the indices $R_1^i$, $R_2^i$, and $R_3^i$ are exclusive integers randomly chosen from the range [1, *Np*], (*Np* is population size in $t^{th}$ generation). *F* is the scaling factor which usually lies in [0, 1] for scaling the difference vectors. $x_{best}^{(t)}$ is the best individual in the poulation at iteration t. In general, mutation strategies can be divided into two categories: exploring biased, exploiting biased and balance mode. For example, the DE/rand/1 belongs to the type of exploring biased, the DE/best/1 belongs to the type of exploiting biased, moreover, DE/current-to-best/1 is balance mode because the diversity is enhanced by adding a difference vector composed of random individuals and the use of best individual. The suitable balance point of exploration and exploitation is changing dynamically in the evolutionary process, thence, an adaptive mutation strategy based on feasible rate is proposed in the paper to switch state between exploring biased and exploiting biased in order to keep the algorithm in an efficient state. In the proposed algorithm (APDE-NS), the mutation strategy will switch to the highly exploiting biased when the current state is unable to find a feasible solution accurately, which will switch back to the initial balance mode status after finding enough feasible solutions. Despite the highly exploiting biased mutation strategy might increase the probability of premature convergence, it is efficient in finding feasible solutions. Therefore, it is suitable to use the highly exploiting biased status at a particular stage of the evolutionary process by adaptive method.

As is mentioned above, the adaptation of *F* and *Cr* is the main research object. Compared to this, relatively fewer works have been allocated to the adaptation of population size [2]. In this part, several adaptive or non-adaptive population size control methods have been proposed and achieved competitive performance. Zamuda and Brest [17] controlled the population size based on increasing number of *FEs*, moreover, a set of novel mutation strategies which depend on population size were adopted. This method shows its superior performance on the CEC2011 real life problems. Tanabe and Fukunaga [18] employed the linear population size reduction to improve the SHADE algorithm, the population size is continually reduced with the increasing of *FEs* and called this variant as L-SHADE. This reduction will stop if the population size reaches the preset minimum (*Np*min). The *FEs* is the current number of function evaluations. L-SHADE ranked 1st at the CEC2014 competition on real parameter single-objective optimization. The population size controlling-method called Linear Population Size Reduction (LPSR) requires only one parameter (initial population size). Another similar approach Dynamic Population Size Reduction (DPSR) [19] reduces the population by half at preset intervals. These non-adaptive methods can increase or decrease (decrease in most cases) the population size according to preset conditions and significantly simplifies the method of controlling population size. Instead of reducing the population size by fixed formula or predetermined path, several adaptive population size techniques were proposed in recent past. Zhu at al. [20] proposed a novel technique which varies the population size in a predetermined range. The algorithm will detect the current state and decrease or increase the current population size according to the feedbacks from previous iterations. The population size will be expanded by introducing more solutions when the algorithm is unable to find a better solution within a preset number of consequent generations. On the contrary, if the algorithm updates the best solution successfully in successive generations, the population size will reduce in order to purge the redundant solutions. Many experiments and studies have proved the effectiveness of the population size control method.

The proposed algorithm APDE-NS will be evaluated by 28 benchmark functions from CEC2017 constrained real parameter optimization. The performance of APDE-NS is better than, or at least competitive to the state-of-the-art constraint optimization algorithms, showing the competitiveness relative to the other excellent algorithm.

The rest of the paper is organized as follows. We review the classical DE algorithm and several constraint handling methods for constrained optimization in Section II, Section III gives the detailed description of the proposed APDE-NS, computational results of benchmark functions and comparison with other state-of-the-art algorithms are shown in Section IV. Finally, Section V summarizes the conclusions and discusses the direction of future work.

## II. CLASSICAL DE ALGORITHM AND CONSTRAINT HANDLING METHODS

### A. Classical DE Algorithm

The evolutionary process of DE consists of three main operators: mutation operation, crossover operation and selection operation. Similar to other evolutionary algorithms for optimization problems, DE is an algorithm based on population, a DE population consists of *Np* individual vectors $x_{i,t} = (x_{1,t}, \dots, x_{D,t}), i = 1, \dots, Np$, where D is the variable dimension of the optimization target, and *Np* is the population size in the current generation, the population individual vectors are generated randomly at the beginning of evolution. *Np* trail vectors are generated from survived population individual by mutation operation and crossover operation in each generation t until some termination criteria are reached. The detailed evolutionary process will be discussed according to the scheme of Storn and Price [6] as follows.

*Mutation Operation*: After initialization, DE adopts the mutation operation which is based on the difference of other individuals to generate mutant vector $V_{i,t}$ with respect to each target vector $x_{i,t}$, in the current generation. The mutation operation used in classical DE is called DE/rand/1/bin, the formula is represented as

"DE/rand/1":
$$v_i^{(t)} = x_{R_1^i}^{(t)} + F\left(x_{R_2^i}^{(t)} - x_{R_3^i}^{(t)}\right) \quad (5)$$

*Crossover Operation*: After the mutation, Crossover operation is used to generate trail vector $u_{i,t}$ by replacing the components between the target vector $x_{i,t}$ and the mutation vector $v_{i,t}$ with probability. Binomial crossover, which is the most frequently employed crossover operator in the DE implemented as follows:

$$u_{i,t}^j = \begin{cases} v_{i,t}^j & if\ rand[0,1) \leq CR_i\ or\ j = j_{rand} \\ x_{i,t}^j & otherwise \end{cases} \quad (6)$$

Where the $j_{rand}$ is a decision variable index selected from [1, D] randomly. D is the dimension of decision variable.

*Selection Operation*: The selection operation based on greedy selection method chooses survived vector from the trail vector and the target vector and can be represented as

$$x_{i,t+1} = \begin{cases} u_{i,t} & if\ f(u_{i,t}) \leq f(x_{i,t}) \\ x_{i,t} & otherwise \end{cases} \quad (7)$$

The selection operation in the classical DE compares each target vector $x_{i,t}$ against corresponding trial vector $u_{i,t}$ based on the objective function value, keeping the evolution continuing.

*B. Constraint Handling Techniques*

In the real world, constraints exist in most optimization problems. Constraints can be classified as inequality constraints and equality constraints according to their features. In order to solve the COPs by using EAs, the constraint-handling methods are necessary [41]. A constrained optimization problem with D variable dimensions is usually represented as a nonlinear programming of the following form [22]

Minimize: $f(x), x = (x_1, x_2, \dots, x_n)\ and\ x \in \psi$
Subject to: $g_i(x) \leq 0, i = 1, \dots, q$
$h_j(x) = 0, j = q + 1, \dots, m \quad (8)$

$\psi$ is the search space, $g_i(x)$ is the inequality constraint, and $h_j(x)$ is the equality constraint. The number of inequality constraints is $q$, $(m-q)$ is the number of the equality constraints. In general, if the global optimum satisfies the constraints $g_i(x) = 0$, the inequality constraints are called active constraints. Similarly, Active constraints contain all the equality constraints. For convenience, we can transform equality constraints into the form of inequality constraints. The inequality constraints and equality constraints can be integrated into a combined form as

$$G_i(x) = \begin{cases} max\{g_i(x), 0\} & i = 1, \dots, p \\ max\{|h_i(x)| - \delta, 0\} & i = p + 1, \dots, m \end{cases} \quad (9)$$

Where $\delta$ is a tolerance parameter which is usually set as 1e-4. The overall constraint violation for a solution is expressed as

$$v(x) = \frac{\sum_{i=1}^{m} G_i(x)}{m} \quad (10)$$

$v(x)$ is zero for feasible solution and positive when at least one constraint is violated [22], and the overall constraint violation is an important parameter in order to guide the evolutionary process towards feasible areas. In this paper, three popular constraint handling techniques will be introduced as follows

*Superiority of Feasible Solution (SF)*: Two individual $x_1$ and $x_2$ are compared in SF [15, 16]. $x_1$ is considered as superior to $x_2$ if the following conditions are satisfied

1) $x_1$ is a feasible solution but $x_2$ is an unfeasible solution.
2) Both of $x_1$ and $x_2$ are feasible solutions and the objective function value of $x_1$ is smaller than $x_2$.
3) Both of $x_1$ and $x_2$ are infeasible solutions, and the overall constraint violation $v(x)$ of $x_1$ is smaller than $x_2$.

*ε-Constraint (EC)*: The method of Constraint violation and $\varepsilon$ level comparisons was proposed in [23]. The core of $\varepsilon$-Constraint handling technique is the balance between the constraint violation and the objective function value. A parameter $\varepsilon$ was employed to adjust the balance, and the constraint violation $\theta(x)$ was defined as a form based on the maximum of all constraint violations.

$$\theta(x) = max\{max\{0, g_1(x), \dots, g_q(x)\}, max|h_{q+1}(x), \dots, h_m(x)|\} \quad (11)$$

Where $q$ is the number of inequality constraints and $(m-q)$ is the number of equality constraints.

In constrained optimization problems, an individual is regarded as infeasible solution and its precedence is low if the constraint violation is greater than 0, This precedence is controlled by the parameter $\varepsilon$. When the target vector and the trail vector are compared based on $\varepsilon$-Constraint method, a set of objective function value and constraint violation $(f(x), \theta(x))$ will be evaluated according to the $\varepsilon$-level precedence. $x_a$ is considered as superior to $x_b$ when the following conditions are reached

$$(f(x_a), \theta(x_a)) \leq_\varepsilon (f(x_b), \theta(x_b))\ if: \begin{cases} f(x_a) \leq f(x_b), & if\ \theta(x_a), \theta(x_b) \leq \varepsilon \\ f(x_a) \leq f(x_b), & if\ \theta(x_a) = \theta(x_b) \\ \theta(x_a) < \theta(x_b), & otherwise \end{cases} \quad (12)$$

The $\leq_\varepsilon$ between the individual $x_a$ and $x_b$ means $x_a$ is superior to $x_b$ according the $\varepsilon$-level precedence. When $\varepsilon$ is set as ∞, the value of objective function is the only criterion for judging superiority of two individuals. The property of the $\varepsilon$-Constraint method is the flexibility. We can adjust the parameter $\varepsilon$ according to the features of optimization problem in order to obtain the suitable balance between lower objective function value and lower constraint violation, however, how to determine the suitable balance point in different problems is also a challenge and a drawback in this method.

*Self-Adaptive Penalty (SP)*: Penalty functions method can keep the efficient performance even after obtaining enough

---
**Algorithm 1: Adaptive population partitioning**
**Start**
1. **Calculate** successful rates of each subpopulation in previous generation
2. **Obtain** the size of each subpopulation as shown in Eq.20
3. **For** each subpopulation $[m_1^G, m_2^G, \ldots, m_k^G]$ **do**
4.  **Assign** individuals to each population randomly based on their sizes obtained in the step 2
5. **End for**
**End**

---

Fig. 1. Pseudo-code of adaptive population partitioning

feasible individuals with simple structure [24]. Static penalty functions method is popular in constrained optimization due to the simplicity. However, a major drawback of static penalty functions method is that it is difficult to choose the suitable penalty coefficients. Despite this, many methods based on penalty functions provide very competitive results [39]. In general, the penalty coefficients are problem-dependent, we need a prior experience of the solved problem to tune the penalty coefficients. Adaptive penalty functions [25] are proposed in order to improve this defect. The penalty coefficients can be adjusted adaptively according to the information gathered from previous iterations in evolution.

In constrained optimization, it is important to extract useful information from infeasible individual whose worth is low but can guide the evolutionary path to feasible region. The main difference is their precedence of distinct types of infeasible individuals [26]. An adaptive penalty functions method based on distance value and two penalties was proposed in [26]. The amount of added penalties is affected by the feasible rate of population individuals. A greater penalty will be added to infeasible individuals if the feasible rate is low. In contrast, if there are enough feasible individuals, then the infeasible individuals with high-fitness values will obtain relatively small penalties. It means that the algorithm can change the status between seeking the optimal solution and finding feasible solutions without presetting parameters. The details of this method is represented as follows

$$F(x) = d(x) + p(x) \quad (13)$$

$$d(x) = \begin{cases} v(x), & \text{if } pfeas = 0 \\ \sqrt{f''(x)^2 + v(x)^2}, & \text{otherwise} \end{cases} \quad (14)$$

$$p(x) = (1 - \text{pfeas}) \times M(x) + \text{pfeas} \times N(x) \quad (15)$$

$$pfeas = \frac{\text{Number of feasible individual}}{\text{population size}} \quad (16)$$

$$M(x) = \begin{cases} 0, & \text{if } pfeas = 0 \\ v(x), & \text{otherwise} \end{cases} \quad (17)$$

$$N(x) = \begin{cases} 0, & \text{if } x \text{ is a feasible solution} \\ f''(x), & \text{if } x \text{ is an infeasible solution} \end{cases} \quad (18)$$

$$f''(x) = \frac{(f(x) - f_{min})}{(f_{max} - f_{min})} \quad (19)$$

Where $pfeas$ is the feasible rate, $v(x)$ is the overall constrain violation, $f''(x)$ is the normalized fitness value, $f_{max}$ and $f_{min}$ are the maximum and minimum values of the objective function in current generation. $p(x)$ is the penalty value. Thus, the individual with lower overall constraint violation and high fitness can beat the feasible individual with low fitness value in some cases.

## III. ADAPTIVE POPULATION SIZE DE WITH NOVEL MUTATION STRATEGY

As we known, DE is a problem-dependent algorithm which has different suitable parameter ranges when dealing with different problems. Except for parameters like *Cr*, *F*, *Np*, mutation strategies employed in evolution have a significant influence on the performance in searching process. In fact, even for a single problem which has been determined its characteristics, the suitable ranges of parameters like *Cr*, *F*, *Np* are changing following the switch of evolutionary stages. In the meanwhile, the requirements for the mutation strategies is changing especially in constrained optimization. To overcome these difficulty, parameter adaptation method for *Cr*, *F* based on Success-History is employed in many algorithm like SaDE [7], SHADE [10], JADE [8], sTDE-dR [27], Population size controlling methods for *Np* are used on L-SHADE [18], sTDE-dR [27]. Adaptive Multipopulation method and competing strategies method are popular due to the greater amount of computation can be assigned to the most suitable strategy. Despite the adaptive method based on success improve the robustness of algorithms and the optimized performance, when solving constrained problems, there are still some drawbacks as follows:

*1)* The parameter adaptation method based on Success-History has update latency, the reason for this update latency is a parameter called Learning-Period which controls the learning generations of the referred value of *Cr* and *F*. In other words, the values of *Cr* and *F* used in the current generation computed by this adaptive method are generated based on the previous information. Therefore, some more rapid phase changes may not employ the appropriate parameters in constrained optimization. This leads to the latency of algorithms switching between exploitation and exploration flexibly.

*2)* In constrained optimization, an important reference indicator in most adaptive parameter control mechanisms commonly not considered. The state which matches algorithm is different after finding the first feasible solution successfully. When the algorithm has not found any feasible solution, the state the algorithm need is the highly exploration biased, on the contrary, the balance state is suitable for the algorithm. When the number of function evaluations (*FEs*) is close to the *FEs*$_{max}$, the exploiting biased status can improve the effect of convergence. In order to utilize the feasible rate ($pfeas$) effectively, an adaptive population size method and a set of novel mutation strategies based on $pfeas$ are proposed in the paper.

The proposed algorithm APDE-NS will be described more detailed in the rest of this section.

### A. Adaptive Population Partitioning Mechanism

For controlling the state switch efficiently, adaptive population partitioning mechanism is employed in the proposed algorithm. The mechanism divides the whole population $M(G) = [x_1^G, x_2^G, \ldots, x_{Np}^G]$ into $k$ subpopulation $[m_1^G, m_2^G, \ldots, m_k^G]$, $m_k^G$ is the $k$[th] subpopulation of $G$ generation. Each subpopulation has own mutation strategy and adaptive parameter archive. The size of subpopulations initializes to equal value in the first generation. In the preceding steps of evolution, the size of the subpopulations will change according to the success rate of evolution. If the trail vector generated from individual $x_{i,G}^s$ is better than its parent in terms of the fitness value, we can consider the evolution process of the $i$[th] individual in $s$[th] subpopulation as a successful one. The pseudo-code of the adaptive population partitioning mechanism is shown in Fig. 1, and the size of each subpopulation is calculated as

$$Sp^s = \frac{Se_s + Se_0}{\sum_{i=1}^{k}(Se_i + Se_0)} \times Np_G \quad (20)$$

where $Se_s$ is the current count of the successful evolutions in $s$[th] subpopulation. $Se_0$ is a small positive constant in order to avoid zero denominator when the algorithm is close to convergence. $Np_G$ is the entire population size in the $G$[th] generation.

**Algorithm 2: Adaptive population size based on *pfeas***
**Start**
1. **If** $EFs < 0.5 \times FEs_{max}$
2.   **Calculate** the population size as Eq.21, and the population size control method enter the exploring state
3. **Else**
4.   **If** $pfeas > 0.5 \&\& EFs < 0.9 \times FEs_{max}$
6.     **Calculate** the population size as Eq.21
7.   **Else**
8.     **record** the $Npfix$ and $FEsfix$
9.   **End if**
10.  **If** $pfeas = 0 \&\& EFs > 0.9 \times FEs_{max}$
11.    **Calculate** the population size as Eq.21
12.  **Else**
13.    **Calculate** the population size as Eq.22
14.  **End if**
15. **End if**
16. **If** $EFs > 0.9 \times FEs_{max} \&\& pesek==0 \&\& Np < Np_{max}/6$ $\&\& pfeas<0.6$
17.  **Set** population size as $Np_{max}/6$ and set pesek as 1
18. **End if**
**End**

Fig. 2. Pseudo-code of the mechanism of adaptive population size

**Algorithm 3: Parameter adaptation based on success-history**
**Start**
1. **For** each subpopulation $[m_1^G, m_2^G, ..., m_k^G]$ **do**
2.   **Calculate** $mean_{WL}(Tp_{CR})$ and $mean_{WA}(Tp_F)$ according to Eq.23-28
3.   **Obtain** $\mu_{CR}$ and $\mu_F$ based on the results of step 2
4.   **Generate** a random index $H$ ($1 \leq H \leq S_{size}$)
5.   **Update** the corresponding archive $S_j$ of current subpopulation according to $H$, $\mu_{CR}$ and $\mu_F$
6. **End** for
7. **For** each subpopulation $[m_1^G, m_2^G, ..., m_k^G]$ **do**
8.   **Generate** a random index c to select $Cr$ and $F$ from $S_j$
9.   **Calculate** $Cr$ and $F$ used in current generation as Eq.29-30
10. **End** for
**End**

Fig. 3. Pseudo-code of the parameter adaptation mechanism

The size of each subpopulation is determined based on its historical success information. It guarantees that the subpopulation with more suitable features obtains the larger population size.

*B. Adaptive Population Size Based On pfeas*

The population size is an important parameter for improving the effectiveness of the algorithm, some population size control techniques showed competitive performance in [18], [28]-[29]. For the purpose of adapting to the special requirements of constrained optimization, an adaptive population size method based on *pfeas* is proposed in this study. The method divides the evolutionary process into three stages namely, exploring biased state, balance search state, and exploiting biased state. The detailed description with respect to these states is present in below

*The exploring biased state*: The algorithm enters the exploring state before the *FEs* exceeds the $0.5*FEs_{max}$. At this stage, the control method of population size is linear reduction. The population size in this stage is computed according to the formula as follows

$$Np(G + 1) = \text{Round}\left[Np_{max} - \left(\frac{Np_{max} - Np_{min}}{FEs_{max}}\right) \times FEs\right] \quad (21)$$

Where $Np_{min}$ is the preset minimum population size, $Np_{max}$ is the initial population size. Before the *FEs* exceeds the $0.5*FEs_{max}$, the population is large enough to keep powerful exploration and population diversity. The minimum of the population size is $0.5*Np_{max}$. It is a suitable choice for the algorithm to enter the exploring state when we do not know whether the constraints are loose.

*The balance search state*: The algorithm will enters the balance search state if no feasible solution is found when *FEs* reaches $0.5*FEs_{max}$. In this state, the population size will keep the same size ($0.5*Np_{max}$) at the end of the previous state to ensure a high-level diversity. In the meanwhile, the mutation strategies will switch to highly exploiting biased state to enhance the search efficiency. Once the first feasible is successfully found, the population size will be adaptively reduced based on the feasible rate (*pfeas*) which can be formulated as

$$NP(G + 1) = \text{Round}\left[Npfix - \frac{(Npfix - Np_{min})}{(FEs_{max} - FEsfix)} \times (FEs - FEsfix) \times pfeas\right] \quad (22)$$

Where $Npfix$ is the population size in the generation when the algorithm enter to the exploiting biased state, $FEsfix$ is the number of function evaluations when the algorithm enter the exploiting biased state. If *pfeas* is 0, the population size will suspend reduction for keeping enough diversity. When *pfeas* exceeds 0.5, the population size will be calculated as Eq.22. If

**Algorithm 4: State-switch DE/current-to-pbest/1**
Start
1. For i=1 to $Sp^s$ do
2.    Generate $CR$ and $F$ using algorithm 3
3.    Select $x_{r1}$, $x_{pbest}$ from $m_n^G$ and select $x_{r2}$ from $m_n^G \cup Fa$
4.    If $feas > 0$
       Choose the balance search state as described in Eq.31
5.    Else if $pfeas = 0$ && $0.08*FEs_{max} < FEs < 0.5*FEs_{max}$ && $rand < 0.12$
6.       Choose the exploitation biased state as given in Eq.32
7.    Else if $pfeas = 0$ && $0.5*FEs_{max} < FEs < 0.9*FEs_{max}$
8.       Choose the highly exploitation biased state as Eq.33
9.    End if
10. Generate the corresponding trail vector $u_i^s$ using crossover operator
11. If $u_i^s(G)$ is better than $x_i^s(G)$
12.    $x_i^s(G+1) = u_i^s(G)$
13. Else
14.    $x_i^s(G+1) = x_i^s(G)$ and insert $u_i^s(G)$ to Fa
15. End if
16. End for
End

Fig. 4. Pseudo-code of State-switch DE/current-to-pbest/1

**Algorithm 5: State-switch DE/randr1/1**
Start
1. For i=1 to $Sp^s$ do
2.    Generate $Cr$ and $F$ using algorithm 3
3.    Select $x_{r1}$, $x_{r2}$ and $x_{r3}$ from $m_s^G$
4.    Rank $x_{r1}$, $x_{r2}$ and $x_{r3}$ based on their suitable value, the best point is $\dot{x}_{r1}$, $\dot{x}_{r2}$ and $\dot{x}_{r3}$ are selected from remaining points.
5.    If $feas = 0$ && $rand<0.5$
       Choose the exploration biased state as described in Eq.35
6.    Else
7.       Choose the balance search state as shown in Eq.36
8.    End if
10. Generate the corresponding trail vector $u_i^s$ using crossover operator
11. If $u_i^s(G)$ is better than $x_i^s(G)$
12.    $x_i^s(G+1) = u_i^s(G)$
13. Else
14.    $x_i^s(G+1) = x_i^s(G)$
15. End if
16. End for
End

Fig. 5. Pseudo-code of State-switch DE/randr1/1

*pfeas* has reached 0.5 before the second stage begins, The method based on linear reduction will be employed continuously.

*The exploiting biased state*: The algorithm will enter the exploiting biased state when *FEs* reaches $0.9*FEs_{max}$. The population linear reduction method will be employed at this stage. The population will initialize as $Np_{max}/6$ if the population size is greater than $Np_{max}/6$ (*pfeas* remains low or no feasible solution can be found in the first two stages). The mutation strategies will switch to balance mode to enhance the effect.

The pseudo-code of the complete population size controlling mechanism is given as Fig.2.

*C. Parameter Adaptation Based On Success-History*

The choice of parameter $F$ and $Cr$ is crucial to the performance of the algorithm. In the APDE-NS, a parameter adaptation method based on success-history is employed to control $F$ and $Cr$. Each subpopulation has an independent archive $S_j$ ($j = 1,2,...,k$) to store suitable means of $Cr$ and $F$ ($\mu_{CR}$ and $\mu_F$). The size of archive $S_{size}$ is a preset constant. The values of $F$ and $Cr$ used by successful individuals in current generation are stored in temporary archives $Tp_{Cr}$ and $Tp_F$ to calculate the $\mu_{Cr}$ and $\mu_F$. The value of $\mu_{Cr,l}$ and $\mu_{F,l}$ ($l = 1,2,...,S_{size}$) are initialized to 0.5. In each generation, An index $H$ ($1 \leq H \leq S_{size}$) is generated to update the archive $S$. The value located in the $H^{th}$ position is replaced by the new $\mu_{CR}$ or $\mu_F$. $H$ is set to 1 at the beginning and incremented when a new element is inserted into the archive $S$. The contents of the mechanism are showed as follows

$$\mu_{CR} = mean_{WL}(Tp_{CR}) \quad if \ Tp_{Cr} \neq \emptyset \tag{23}$$

$$\mu_F = mean_{WA}(Tp_F) \quad if \ Tp_F \neq \emptyset \tag{24}$$

$$mean_{WA}(Tp_{Cr}) = \sum_{m=1}^{|Tp_{Cr}|} w_m \, Tp_{Cr,m} \tag{25}$$

$$mean_{WA}(Tp_F) = \sum_{m=1}^{|Tp_F|} w_m \, Tp_{F,m} \tag{26}$$

$$w_m = \frac{f_m}{\sum_{i=1}^{|Tp_F|} f_i} \tag{27}$$

$$f_m = |f(x_m) - f(u_m)| \tag{28}$$

Where $f(x_m)$ is the objective function value of the individual which employs the $m^{th}$ $F$ stored in $Tp_F$. $f(u_m)$ is the objective value of the trail vector of the individual. The pseudo-code of the parameter adaptation mechanism is shown as Fig.3.

When the population partitioning is completed, each individual in subpopulations selects an associated $F$ and $Cr$ from the corresponding archive $S_j$. A random index $c \in [1, S_{size}]$ is generated to assign the $Cr$ and $F$ stored in the location to the individual. The final values of $Cr$ and $F$ are calculated as follows

$$Cr = randn(\mu_{CR}, 0.1) \tag{29}$$
$$F = randn(\mu_F, 0.1) \tag{30}$$

The final values of $Cr$ and $F$ are generated based on normal distributions.

*D. A Set of Novel Mutation Strategies for Constrained Optimization*

In constrained optimization, the algorithm's requirements for search characteristics will change after enough feasible solutions have been found. However, most mutation strategies are employed in a fixed manner during evolution and it makes sense to develop novel mutation strategies [21]. A set of novel mutation strategies are proposed for solving constrained problems efficiently. Each subpopulation has a different mutation strategy. These mutation strategies will be described as follows.

TABLE I. ALGORITHM COMPUTATIONAL COMPLEXITY

|        | T1    | T2    | (T2 − T1)/T1 |
|--------|-------|-------|--------------|
| D = 10 | 0.71s | 1.26s | 0.77         |
| D = 30 | 0.74s | 1.49s | 1.01         |
| D = 50 | 0.76s | 2.21s | 1.91         |

*State-switch DE/current-to-pbest/1*: The prototype of our proposed mutation strategy is the DE/current-to-pbest/1/bin introduced in JADE algorithm. The State-switch DE/current-to-pbest/1 has three states: highly exploitation biased state, exploitation biased state and balance search state. The state is switched based on $pfeas$ and $FEs$.

Fig. 4 shows the pseudo-code of the state-switch DE/current-to-pbest/1/bin strategies. The specific formula of each state is represented as follows

*1)* The balance search state:

$$v = x_i + F \cdot (x_{pbest} - x_i) + F \cdot (x_{r1} - x_{r2})\ if\ feas > 0 \quad (31)$$

*2)* The exploitation biased state:

$$v = x_i + F \cdot (x_{pbest} - x_i)$$
$$if\ \begin{cases} pfeas = 0 \\ 0.08 * FEs_{max} < FEs < 0.5 * FEs_{max} \\ rand < 0.12 \end{cases} \quad (32)$$

*3)* The highly exploitation biased state:

$$v = x_i + F \cdot (x_{pbest} - x_i)$$
$$if\ \begin{cases} pfeas = 0 \\ 0.5 * FEs_{max} < FEs < 0.9 * FEs_{max} \end{cases} \quad (33)$$

Where $x_{pbest}$ is one of the top-p individuals ranked according to the overall constraint violations and objective function values. $x_i$ is the current individual, $x_{r1}$ is selected from the subpopulation $m_n^G$ which contains $x_i$. $x_{r2}$ is randomly selected from $m_n^G \cup Fa$. Fa is an archive to store individuals failed in evolution. The size of Fa is defined as $Fa_{size} = 2 * Np$. Randomly selected elements in Fa are deleted for newly inserted elements. When neither of the three conditions are satisfied, the formula of the strategy is $v = x_i + F \cdot (x_{pbest} - x_i) + F \cdot (x_{r1} - x_{r2})$.

*State-switch DE/randr1/1*: The prototype of our proposed mutation strategy is DE/randr1/1 proposed in [30], the mutation strategy is a modified version based on popular rand/1. The details in regard to the rand/1 can be obtained in [30]. It has been proved that the DE/randr1/1 mutation strategy can significantly increase search speed compared to the classical rand/1. A further modification on DE/randr1/1 is made in the paper to enhance the efficiency of optimizing constrained problems and a state-switch mechanism is employed in the State-switch DE/randr1/1 strategy. The pseudo-code of State-switch DE/randr1/1 is showed in Fig. 5.

The State-switch DE/randr1/1 strategy has two states: exploration biased state and balance search state. The state is switched based on *pfeas* and diversity index (*rdiv*) calculated as (34). The detailed description of each strategy is showed as follows

$$rdiv_G = \frac{rdivf_G}{rdivf_{init}} \quad (34a)$$

$$rdivf_G = \sqrt{\frac{\sum_{i=1}^{Np_G}\sum_{j=1}^{D} x_{i,j}^G - \overline{x_j^G}}{Np_G}} \quad (34b)$$

Where $Np_G$ is the population size in the $G^{th}$ generation, $D$ is the dimension of variable. $\overline{x_j^G}$ is the mean of $j^{th}$ component in the $G^{th}$ generation. $rdivf_{init}$ is the diverity of the initial population. $rdiv_G$ is 1 at the beginning and continues to decline during the evolution. Therefore, a small *rdiv* corresponds to a poor diversity. When *rdiv* reaches 0, the algorithm can be regarded as convergence.

*1)* The exploration biased state:
$$v = \dot{x}_{r1} + F \cdot (\dot{x}_{r2} - \dot{x}_{r3}) + cc \cdot rdiv \cdot (\dot{x}_{r1} - \dot{x}_{r2})$$
$$if\ pfeas = 0\ \&\&\ rand < 0.5 \quad (35)$$

*2)* The balance search state:

$$v = \dot{x}_{r1} + F \cdot (\dot{x}_{r2} - \dot{x}_{r3})$$
$$else\ the\ conditons\ in\ (35) \quad (36)$$

$x_{r1}$, $x_{r2}$ and $x_{r3}$ are randomly selected from current population, points $x_{r1}$, $x_{r2}$ and $x_{r3}$ are distinct and not equal to $x_i$. $\dot{x}_{r1}$ is the tournament best among the three points, $\dot{x}_{r2}$ and $\dot{x}_{r3}$ is remaining points. $cc$ is a preset constant called influence coefficient.

*E. Choice of Mutation Strategies for Subpopulation*

In the proposed APDE-NS algorithm, the entire population is divided into four subpopulations. Each subpopulation, which is assigned a distinct strategy and a set of novel mutation strategies are utilized as follows

*1)* State-switch DE/current-to-pbest/1/bin
*2)* State-switch DE/current-to-pbest/1/exp
*3)* State-switch DE/randr1/1/bin
*4)* State-switch DE/randr1/1/exp

Where *bin* is crossover operator binomial, and exp is crossover operator exponential. The detailed information with regard to the above two crossover operators can be found in [31]. Whenever $FEs$ reaches $0.9 * FEs_{max}$, the subpopulation employing the third and fourth strategies should be turned to employ the first and the second strategies.

*F. Modified algorithm of APDE-NS for low-dimensional problems*

The APDE-NS algorithm has a competitive performance on high-dimensional problems. However, when solving low-dimensional problems, its performance is not good enough. The inefficiency on low-dimensional problems is caused by two major reasons. Detailed discussion respecting the reasons is as follows

*1)* The delay of parameters (*F*, *Cr*) adaptive mechanism

The parameters adaptive mechanism of *F* and *Cr* is based on success-history, this means $F$ and $Cr$ employed in next generation are calculated from the $F$ and $Cr$ of successful individuals in past generations. Simultaneously, the archive $S$ is only updated one location per generation. The suitable ranges of $F$ and $Cr$ are changing in evolutionary process. In low-dimensional problems, the space complexity is low, the changes of suitable range are not active as changes in high-dimensional problems. Therefore, the negative impact of this delay on low-dimensional problems will be more serious than that on high-dimensional problems.

TABLE II. SUCCESSFUL OPTIMIZATION RATE FOR BENCHMARK FUNCTIONS

| Dimension | 10D | | | | | | | |
|---|---|---|---|---|---|---|---|---|
| Algorithm | APDE-NS-L | CAL-SHADE | L-SHADE44 | SaDE | SajDE | DEbin | L-S44+IDE | UDE |
| SR | 78.57% | 71.42% | 71.42% | 64.29% | 35.71% | 67.86% | 75% | 75% |
| Failure(*f*) | 6 | 8 | 8 | 10 | 18 | 9 | 7 | 7 |
| Dimension | 30D | | | | | | | |
| Algorithm | APDE-NS | CAL-SHADE | L-SHADE44 | SaDE | SajDE | DEbin | L-S44+IDE | UDE |
| SR | 78.57% | 71.42% | 75% | 60.71% | 32.14% | 67.86% | 75% | 75% |
| Failure(*f*) | 6 | 8 | 7 | 11 | 19 | 9 | 7 | 7 |
| Dimension | 50D | | | | | | | |
| Algorithm | APDE-NS | CAL-SHADE | L-SHADE44 | SaDE | SajDE | DEbin | L-S44+IDE | UDE |
| SR | 78.57% | 67.86% | 75% | 57.14% | 28.57% | 60.71% | 75% | 75% |
| Failure(*f*) | 6 | 9 | 7 | 12 | 20 | 11 | 7 | 7 |

*2)* The negative effects of highly constraint control state

In our proposed mutation strategies, the highly exploitation biased state is efficient for control constraints. However, in the low-dimensional problems, constraint control is easier than that of high-dimensional problems, The mechanism of adaptive population size based on *pfeas* and other constraint control mechanism employed in the APDE-NS are enough to control constraints. State switching used in mutation strategies becomes a bit redundant. Keeping a balance search state is better in low-dimensional problems.

In order to improve the effectiveness of the algorithm in low-dimensional problems. The parameter adaptive mechanism of $F$ and $Cr$ and mutation strategies is modified, termed as APDE-NS-L. A parameter pool $P$ replaces the original mechanism based on success-history. Each subpopulation will update the probability of choosing different parameter pair in the pool. The probability of choosing parameter pair evaluated over the previous learning period $LP$. The mutation strategies used in four subpopulation are modified as follows

*a)* $v = x_i + F \cdot (x_{pbest} - x_i) + cc \cdot (1 - rdiv) \cdot (x_{r1} - x_{r2})$
*b)* DE/current-to-pbest/1/bin
*c)* State-switch DE/current-to-pbest/1/bin
*d)* DE/randr1/1/bin

IV. EXPERIMENTAL ANALYSIS

The proposed APDE-NS is tested on 28 widely used benchmark functions from the IEEE CEC2017 benchmarks on constrained real-parameter optimization [32] on 10-D, 30-D and 50-D. More details about these functions can be found in [32], the algorithm was run 25 times for each test function with a number of function evaluations equals to $20000 \times D$ according to the guidelines of CEC2017 [32]. The evaluation criteria employed to evaluate the performance of APDE-NS is the method described in [32] and includes feasibility rate of all runs ($FR$), mean violation amounts ($\overline{vio}$) and mean of objective function value (*mean*). the priority level of the evaluation criteria is as follows [32]

① First**,** evaluate the algorithms based on feasible rate.

② Then, evaluate the algorithm according to the mean violation amounts.

③ At last, evaluate the algorithm in terms of the mean objective function value.

When different algorithm are compared under the same test function, we first compare the feasibility rate (FR) of them. Higher feasibility rate means better performance. If the feasibility rate of them is equal, the algorithm with smaller mean violation is better. If both feasibility rate and mean violation are equal, the algorithm with smaller objective function value is better. Pairwise comparisons method the sign test [38] is also employed to compare APDE-NS with other state-of-the-art algorithm.

*A. Parameters Setting*

The pre-set parameters is showed as follows
*1)* The initial population size $Np_{max}$ was set to 24*D in 10-D problems, 19*D in 30-D and 16*D in 50-D problems.
*2)* The minimum population size $Np_{min}$ was set to 6.
*3)* The number of subpopulation was set to 4.
*4)* The size of archive $S_j$ $(j = 1,2,...,k)$ was set to 6.
*5)* The influence coefficient $cc$ was set to 0.3.
*6)* The parameter pool: [0.9 0.9; 0.5 0.5; 0.9 0.2; 0.6 0.8]

*B. Algorithm Computational Complexity*

The algorithm was run 25 times for each benchmark function with a number of function evaluations equals to 20000*D. The computational complexity of APDE-NS is calculated as the method described in [32] and is shown in Table I. T1 is the computing time of 10000 evaluations for all benchmark functions. T2 is the complete computing time for the algorithm with 10000 evaluations for all benchmark functions, and $(T2 - T1)/T1$ is the algorithm computational complexity.

*C. Comparisons with State-of-Art Algorithms*

In order to evaluate the performance of the proposed APDE-NS algorithm, the following state-of-the-art algorithms including the top four algorithms for constrained optimization in CEC2017 are compared with the APDE-NS.

*1)* Adaptive constraint handling and success history

TABLE III. EXPERIMENTAL RESULTS IN MEAN, FR AND $\overline{VIO}$ ON FUNCTIONS 1-28. 10-D

| Func | APDE-NS-L | CAL-SHADE | L-SHADE44 | SaDE | SajDE | DEbin | L-S44+IDE | UDE |
|---|---|---|---|---|---|---|---|---|
| F1 | **0** | 6.43e-30 | **0** | **0** | 6.18e-30 | 2.72e-29 | **0** | 5.03e-15 |
|  | **1** | 1 | **1** | **1** | 1 | 1 | **1** | 1 |
|  | **0** | 0 | **0** | **0** | 0 | 0 | **0** | 0 |
| F2 | **0** | 9.10e-17 | **0** | **0** | 2.69e-29 | 1.21e-28 | **0** | 6.44e-15 |
|  | **1** | 1 | **1** | **1** | 1 | 1 | **1** | 1 |
|  | **0** | 0 | **0** | **0** | 0 | 0 | **0** | 0 |
| F3 | 26340.2 | 119673 | 81998.4 | 104988 | **1.19e-29** | 103441 | 3.26e+05 | 77.3859 |
|  | 1 | 0.48 | 0.92 | 0.08 | **1** | 0.28 | 1 | 0.96 |
|  | 0 | 6.26e-05 | 9.26e-06 | 0.00019 | **0** | 0.000126 | 0 | 4.30e-06 |
| F4 | 13.7605 | 13.8012 | 13.6285 | **13.5728** | 0.1194 | 14.7929 | 14.42 | 25.1168 |
|  | 1 | 1 | 1 | **1** | 1 | 1 | 1 | 1 |
|  | 0 | 0 | 0 | **0** | 0.0501 | 0 | 0 | 0 |
| F5 | **0** | 0.3189 | **0** | **0** | 0.3190 | 1.2541 | **0** | 1.6827 |
|  | **1** | 1 | **1** | **1** | 1 | 1 | **1** | 1 |
|  | **0** | 0 | **0** | **0** | 0 | 0 | **0** | 0 |
| F6 | 167.412 | 882.993 | 720.49 | 795.119 | **0** | 428.253 | 808.3605 | 8.71e+01 |
|  | 1 | 0 | 0.04 | 0 | **1** | 0.16 | 0 | 0.44 |
|  | 0 | 0.1462 | 0.03195 | 0.1827 | **0** | 0.1418 | 0.03766 | 0.0204 |
| F7 | **-25.4179** | 17.9934 | -6.5994 | 21.918 | -522.473 | 30.6191 | -34.0032 | -6.4624 |
|  | **1** | 0.68 | 0.76 | 0 | 0 | 0.08 | 0.8 | 0.72 |
|  | **0** | 0.0001 | 3.25e-05 | 0.1144 | 1021.63 | 0.001215 | 3.19e-05 | 0.07061 |
| F8 | **-0.00135** | **-0.00135** | **-0.00135** | **-0.00135** | -90.3668 | -0.00029 | **-0.00135** | -0.00134 |
|  | **1** | **1** | **1** | **1** | 0 | 1 | **1** | 1 |
|  | **0** | **0** | **0** | **0** | 476500 | 0 | **0** | 0 |
| F9 | **-0.00498** | 1.0032 | **-0.00498** | **-0.00498** | -0.6092 | 0.0991 | **-0.00498** | **-0.00498** |
|  | **1** | 0.52 | **1** | **1** | 0 | 1 | **1** | **1** |
|  | **0** | 0.0409 | **0** | **0** | 253.304 | 0 | **0** | **0** |
| F10 | **-0.00051** | **-0.00051** | **-0.00051** | **-0.00051** | -59.754 | -0.00045 | **-0.00051** | **-0.00051** |
|  | **1** | **1** | **1** | **1** | 0 | 1 | **1** | 1 |
|  | **0** | **0** | **0** | **0** | 1.11e+06 | 0 | **0** | 0 |
| F11 | -0.1525 | -0.1595 | **-0.1688** | 0.0857 | -986.103 | -0.06452 | **-0.1688** | -5.9955 |
|  | 1 | 0.88 | **1** | 0.68 | 0 | 0.88 | **1** | 1 |
|  | 0 | 0.00018 | **0** | 1.85e-07 | 4.25e+19 | 7.72e-15 | **0** | 3.00e-03 |
| F12 | 3.9880 | 4.0037 | 3.9932 | 3.9934 | 0.1990 | **3.9879** | **3.9879** | **3.9879** |
|  | 1 | 1 | 1 | 1 | 0 | **1** | **1** | **1** |
|  | 0 | 0 | 0 | 0 | 1.9005 | **0** | **0** | **0** |
| F13 | **0** | 0.1867 | 1.97e-31 | **0** | 0.4784 | 0.02578 | **0** | 11.1236 |
|  | **1** | 1 | 1 | **1** | 1 | 1 | **1** | 1 |
|  | **0** | 0 | 0 | **0** | 0 | 0 | **0** | 0 |
| F14 | 2.85696 | 3.2745 | 2.8391 | 2.8478 | 1.42e-16 | 3.1894 | 3.00001 | **2.7362** |
|  | 1 | 1 | 1 | 1 | 0 | 1 | 1 | **1** |
|  | 0 | 0 | 0 | 0 | 2.5 | 0 | 0 | **0** |
| F15 | 13.6659 | 14.925 | 14.2941 | 12.5349 | 2.20e-15 | 14.7968 | **11.2783** | 6.7549 |
|  | 1 | 0 | 0.28 | 0.92 | 0 | 1 | **1** | 0.92 |
|  | 0 | 0.0047 | 0.000189 | 4.30e-06 | 0.5 | 0 | **0** | 0.00037 |
| F16 | 37.4477 | 49.0088 | 40.9035 | 51.2708 | **0** | 47.7522 | 40.4009 | 6.2830 |
|  | 1 | 0.96 | 1 | 1 | **1** | 1 | 1 | 0.96 |
|  | 0 | 2.09e-06 | 0 | 0 | **0** | 0 | 0 | 1.11e-05 |
| F17 | 1.05208 | 0.9213 | 0.91933 | **0.9152** | 0.00128 | 1.0273 | 0.8824 | 1.0460 |
|  | 0 | 0 | 0 | **0** | 0 | 0 | 0 | 0 |
|  | 4.66 | 5.5 | 5.26 | **4.5** | 22.9388 | 4.94 | 5.22 | 5.42 |
| F18 | 382.044 | 797.597 | 3048.86 | 774.891 | **0.08** | 625.308 | 3166.311 | 2359.1321 |
|  | 0 | 0 | 0 | 0 | **0** | 0 | 0 | 0 |
|  | 13542.5 | 115.852 | 1.71e+06 | 57.4051 | **7.4797** | 114.367 | 8.23e+06 | 1.51e+07 |
| F19 | **0** | 2.39e-06 | 1.58e-06 | **0** | -9.2521 | 1.12e-06 | **0** | 0.00272 |
|  | **0** | 0 | 0 | **0** | 0 | 0 | **0** | 0 |
|  | **6633.59** | 6633.59 | 6633.59 | **6633.59** | 6646.37 | 6633.59 | **6633.59** | 6633.59 |
| F20 | 0.4511 | 0.2490 | **0.1801** | 0.4580 | 0.6089 | 0.4279 | 0.4156 | 1.6464 |
|  | 1 | 1 | **1** | 1 | 1 | 1 | 1 | 1 |
|  | 0 | 0 | **0** | 0 | 0 | 0 | 0 | 0 |
| F21 | **3.9879** | 3.9901 | 3.9884 | 3.988 | 10.7765 | **3.9879** | **3.9879** | 6.2436 |
|  | **1** | 1 | 1 | 1 | 0 | **1** | **1** | 1 |
|  | **0** | 0 | 0 | 0 | 1.2066 | **0** | **0** | 0 |
| F22 | 0.3186 | 110.109 | 0.3189 | 0.4784 | 19.3128 | 8.3719 | **0.1595** | 12.5042 |
|  | 1 | 1 | 1 | 1 | 0.84 | 1 | **1** | 1 |
|  | 0 | 0 | 0 | 0 | 237.55 | 0 | **0** | 0 |
| F23 | 2.70887 | 3.3359 | **2.5320** | 3.3042 | 20.4959 | 3.0997 | 3.0212 | 2.7456 |
|  | 1 | 1 | **1** | 1 | 0 | 1 | 1 | 1 |
|  | 0 | 0 | **0** | 0 | 40873.7 | 0 | 0 | 0 |
| F24 | 13.2889 | 10.0216 | **8.1367** | 20.4519 | 1.41e-16 | 11.7809 | 8.7650 | 6.0006 |
|  | 1 | 1 | **1** | 0 | 0 | 1 | 1 | 0.96 |
|  | 0 | 0 | **0** | 111.955 | 0.5 | 0 | 0 | 0.000152 |
| F25 | 46.2441 | 46.3070 | 37.0080 | 68.3611 | **7.15e-15** | 47.6265 | 37.6991 | 6.3459 |
|  | 1 | 1 | 1 | 0.92 | **1** | 1 | 1 | 1 |
|  | 0 | 0 | 0 | 8.11e-06 | **0** | 0 | 0 | 0 |
| F26 | 1.05465 | 0.9509 | 1.0059 | **1.1173** | 0.02058 | 1.0636 | 0.9374 | 1.0217 |
|  | 0 | 0 | 0 | **0** | 0 | 0 | 0 | 0 |
|  | 4.7 | 5.46 | 5.1 | **4.5** | 29.8164 | 4.9 | 5.0526 | 5.4611 |
| F27 | 438.689 | 3176.41 | 2607.86 | **731.712** | 31.8064 | 5860.64 | 7944.1523 | 6717.898 |
|  | 0 | 0 | 0 | **0** | 0 | 0 | 0 | 0 |
|  | 21841.5 | 86412.9 | 357097 | **177.42** | 1858.54 | 264709 | 4.05e+07 | 2.23e+08 |
| F28 | 21.1034 | 35.9246 | 25.7495 | **10.4185** | -0.08029 | 23.984 | 10.7630 | 9.7567 |
|  | 0 | 0 | 0 | **0** | 0 | 0 | 0 | 0 |
|  | 6647.99 | 6660.36 | 6652.22 | **6639.56** | 6647.53 | 6649.59 | 6640.886 | 6640.993 |
| +/=/− Criterion I |  | 23/2/3 | 15/6/7 | 13/8/7 | 21/0/7 | 23/1/4 | 11/9/8 | 22/2/4 |
| +/=/− Criterion II |  | 18/8/2 | 7/17/4 | 11/14/3 | 19/3/6 | 15/12/1 | NA | NA |

differential evolution for CEC2017 constrained real-parameter optimization (CAL-SHADE) [33].

*2)* A simple framework for constrained problems with application of L-SHADE44 and IDE (L-SHADE44+IDE) [34].

*3)* L-SHADE with competing strategies applied to constrained optimization (L-SHADE44) [35].

TABLE IV. EXPERIMENTAL RESULTS IN MEAN, FR AND $\overline{vio}$ ON FUNCTIONS 1-28. 30-D

| Func | APDE-NS | CAL-SHADE | L-SHADE44 | SaDE | SajDE | DEbin | L-S44+IDE | UDE |
|---|---|---|---|---|---|---|---|---|
| F1 | 5.75e-30 | 9.63e-29 | 7.23e-30 | 1.16e-07 | 4.69e-27 | 1.01e-11 | **0** | 7.34e-29 |
|  | 1 | 1 | 1 | 1 | 1 | 1 | **1** | 1 |
|  | 0 | 0 | 0 | 0 | 0 | 0 | **0** | 0 |
| F2 | 2.67e-30 | 8.66e-29 | 4.74e-30 | 6.31e-08 | 1.17e-26 | 7.17e-12 | **0** | 7.39e-29 |
|  | 1 | 1 | 1 | 1 | 1 | 1 | **1** | 1 |
|  | 0 | 0 | 0 | 0 | 0 | 0 | **0** | 0 |
| F3 | 342799 | 369122 | 317283 | 960694 | **7.42e-27** | 1.16e+06 | 6.70e+06 | 73.254 |
|  | 1 | 1 | 1 | 0.44 | **1** | 0.44 | 1 | 1 |
|  | 0 | 0 | 0 | 0.0001 | **0** | 0.00012 | 0 | 0 |
| F4 | **13.5728** | 13.8334 | **13.5728** | 13.5729 | 0.3582 | 13.6667 | 13.8544 | 82.4219 |
|  | **1** | 1 | **1** | 1 | 1 | 1 | 1 | 1 |
|  | **0** | 0 | **0** | 0 | 0.1502 | 0 | 0 | 0 |
| F5 | **0** | 0.1595 | 3.65e-31 | 1.47e-07 | 5.4837 | 1.01567 | **0** | 2.32e-17 |
|  | **1** | 1 | 1 | 1 | 1 | 1 | **1** | 1 |
|  | **0** | 0 | 0 | 0 | 0 | 0 | **0** | 0 |
| F6 | 4453.81 | 3910.24 | 3733.02 | 4262.81 | 0.4776 | 3729.59 | 5526.359 | **303.62** |
|  | 0.32 | 0 | 0 | 0 | 0.56 | 0 | 0 | **1** |
|  | 0.01306 | 0.0208 | 0.01780 | 0.2925 | 0.2914 | 0.3108 | 0.02567 | **0** |
| F7 | -93.038 | -150.464 | -108.879 | -35.9309 | -1546 | 25.5352 | -81.0877 | **-598.1343** |
|  | 1 | 1 | 0.96 | 0.12 | 0 | 0.16 | 0.96 | **1** |
|  | 0 | 0 | 5.51e-06 | 0.0011 | 2905.53 | 0.00168 | 4.06e-06 | **0** |
| F8 | **-0.00028** | **-0.00028** | **-0.00028** | -0.00018 | -63.808 | -0.00021 | -0.00026 | **-0.00028** |
|  | **1** | **1** | **1** | 1 | 0 | 1 | 1 | **1** |
|  | **0** | **0** | **0** | 0 | 7.80e+06 | 0 | 0 | **0** |
| F9 | **-0.00267** | **-0.00267** | **-0.00267** | **-0.00267** | -0.5871 | 0.03425 | **-0.00267** | **-0.00267** |
|  | **1** | **1** | **1** | **1** | 0 | 0 | **1** | **1** |
|  | **0** | **0** | **0** | **0** | 12748.6 | 0 | **0** | **0** |
| F10 | **-0.0001** | **-0.0001** | **-0.0001** | -9.98e-05 | -50.4385 | **-0.0001** | -9.80e-05 | **-0.0001** |
|  | **1** | **1** | **1** | 1 | 0 | **1** | 1 | **1** |
|  | **0** | **0** | **0** | 0 | 2.95e+07 | **0** | 0 | **0** |
| F11 | -0.8536 | -0.4539 | -0.8623 | 1.9375 | -2968.71 | -20.5126 | **-0.8651** | -28.348 |
|  | 1 | 0.88 | 1 | 0 | 0 | 0 | **1** | 0 |
|  | 0 | 1.0e-19 | 0 | 1.16e+22 | 3.47e+59 | 1.5547 | **0** | 0.0207 |
| F12 | 3.9856 | 21.1864 | 3.9856 | 3.9963 | 0.9950 | **3.9826** | 6.0679 | 18.679 |
|  | 1 | 1 | 1 | 1 | 0 | **1** | 1 | 1 |
|  | 0 | 0 | 0 | 0 | 1.6219 | **0** | 0 | 0 |
| F13 | 6.9248 | 15.0401 | **4.8409** | 13.4271 | 5.9520 | 23.1947 | 26.0108 | 81.4897 |
|  | 1 | 1 | **1** | 1 | 1 | 1 | 1 | 1 |
|  | 0 | 0 | **0** | 0 | 0 | 0 | 0 | 0 |
| F14 | 1.8808 | 2.0530 | 1.8330 | 1.8373 | 0.03725 | 2.1033 | 1.9086 | **1.5258** |
|  | 1 | 1 | 1 | 1 | 0 | 1 | 1 | **1** |
|  | 0 | 0 | 0 | 0 | 2.4701 | 0 | 0 | **0** |
| F15 | 18.6925 | 19.4466 | 18.8181 | 16.3049 | 15.3211 | 18.5668 | **12.9119** | 9.142 |
|  | 0.96 | 0.16 | 0.84 | 1 | 0 | 1 | **1** | 0.88 |
|  | 3.23e-06 | 0.00043 | 1.86e-05 | 0 | 357.848 | 0 | **0** | 2.62e-05 |
| F16 | 146.524 | 160.724 | 148.534 | 194.339 | **4.97e-16** | 178.505 | 143.0053 | 8.4193 |
|  | 1 | 1 | 1 | 1 | **1** | 1 | 1 | 1 |
|  | 0 | 0 | 0 | 0 | **0** | 0 | 0 | 0 |
| F17 | 1.0019 | 1.0141 | **0.9984** | 1.0221 | 0.00256 | 1.02634 | 1.0128 | 1.0214 |
|  | 0 | 0 | **0** | 0 | 0 | 0 | 0 | 0 |
|  | 15.5 | 15.5 | **15.5** | 15.5 | 70.3902 | 15.5 | 15.5 | 15.5 |
| F18 | 827.291 | 4831.74 | 1583.02 | 589.272 | **0.68** | 6170.07 | 4009.6169 | 3568.1866 |
|  | 0 | 0 | 0 | 0 | **0** | 0 | 0 | 0 |
|  | 253662 | 1.11e+06 | 15793.5 | 2052.77 | **47.2879** | 2.04e+06 | 1.33e+06 | 3.36e+07 |
| F19 | 5.35e-06 | 7.33e-06 | 6.29e-06 | **0** | -27.7475 | 7.81e-06 | 0 | 3.6576 |
|  | 0 | 0 | 0 | **0** | 0 | 0 | 0 | 0 |
|  | 21374.9 | 21374.9 | 21374.9 | **21374.9** | 21416 | 21374.9 | 21374.9 | 21378.9 |
| F20 | 1.4602 | **1.1764** | 1.4034 | 3.0345 | 1.8854 | 3.1429 | 2.2476 | 4.5816 |
|  | 1 | **1** | 1 | 1 | 0.96 | 1 | 1 | 1 |
|  | 0 | **0** | 0 | 0 | 0.00331 | 0 | 0 | 0 |
| F21 | 22.9321 | 19.4655 | 18.6949 | **9.5185** | 85.7652 | 18.2165 | 27.4671 | 11.9103 |
|  | 1 | 1 | 1 | **1** | 0 | 1 | 1 | 0.92 |
|  | 0 | 0 | 0 | **0** | 40.6659 | 0 | 0 | 1.48e-12 |
| F22 | 213.539 | 33571.2 | 3064.06 | 26707.3 | 121.355 | 3669.25 | 844.0106 | **94.1472** |
|  | 1 | 0.88 | 1 | 0.36 | 0.72 | 0.92 | 1 | **1** |
|  | 0 | 0.4042 | 0 | 3.2106 | 291.938 | 0.05175 | 0 | **0** |
| F23 | 1.7837 | 2.0591 | 1.6560 | 1.9966 | 21.0515 | 2.0072 | 1.8564 | **1.4349** |
|  | 1 | 1 | 1 | 1 | 0 | 1 | 1 | **1** |
|  | 0 | 0 | 0 | 0 | 252239 | 0 | 0 | **0** |
| F24 | 12.0322 | 15.0482 | 12.1579 | 59.0278 | 1.4898 | 17.0588 | 13.9172 | **8.3879** |
|  | 1 | 1 | 1 | 0 | 0 | 1 | 1 | **1** |
|  | 0 | 0 | 0 | 8652.35 | 0.5315 | 0 | 0 | **0** |
| F25 | 139.361 | 166.441 | 141.434 | 318.937 | 14.7717 | 177.123 | 140.6177 | **15.9592** |
|  | 1 | 1 | 1 | 0 | 0 | 1 | 1 | **1** |
|  | 0 | 0 | 0 | 1151.98 | 0.6795 | 0 | 0 | **0** |
| F26 | 1.0062 | 1.0176 | **1.0056** | 1.0264 | 0.008572 | 1.0277 | 1.0123 | 1.0194 |
|  | 0 | 0 | **0** | 0 | 0 | 0 | 0 | 0 |
|  | 15.5 | 15.5 | **15.5** | 15.5 | 58.3465 | 15.5 | 15.5 | 15.5 |
| F27 | 2337.63 | 12675.4 | 9544.44 | **500.718** | 198.896 | 24305.3 | 52293.45 | 11876.39 |
|  | 0 | 0 | 0 | **0** | 0 | 0 | 0 | 0 |
|  | 4.98e+06 | 3.83e+07 | 1.00e+07 | **16451.1** | 57097.8 | 2.08e+08 | 1.08e+08 | 3.41e+08 |
| F28 | 111.853 | 159.71 | 136.058 | 146.273 | 38.1999 | 108.069 | 152.8521 | **64.323** |
|  | 0 | 0 | 0 | 0 | 0 | 0 | 0 | **0** |
|  | 21459.4 | 21492.3 | 21477.6 | 21491.2 | 21465.6 | 21462 | 21481.66 | **21446.29** |
| +/=/− Criterion I |  | 22/3/3 | 13/5/10 | 20/1/7 | 22/0/6 | 24/1/3 | 20/2/6 | 16/3/9 |
| +/=/− Criterion II |  | 16/11/1 | 4/21/3 | 20/5/3 | 20/4/4 | 21/5/2 | NA | NA |

*4)* A unified differential evolution algorithm for constrained optimization problems (UDE) [36].

*5)* Self-adaptive differential evolution algorithm for constrained real-parameter optimization (SaDE) [37].

*6)* An Improved Self-adaptive Differential Evolution Algorithm in Single Objective Constrained Real-Parameter

# > REPLACE THIS LINE WITH YOUR PAPER IDENTIFICATION NUMBER (DOUBLE-CLICK HERE TO EDIT) < 11TABLE V. Experimental Results in mean, fr and $\overline{vio}$ on Functions 1-28. 50-D

| Func | APDE-NS | CAL-SHADE | L-SHADE44 | SaDE | SajDE | DEbin | L-S44+IDE | UDE |
|---|---|---|---|---|---|---|---|---|
| F1 | 8.57e-29 | 6.38e-28 | **7.75e-29** | 0.02424 | 5.10e-15 | 0.1425 | 1.49e-08 | 1.16e-10 |
|  | 1 | 1 | **1** | 1 | 1 | 1 | 1 | 1 |
|  | 0 | 0 | **0** | 0 | 0 | 0 | 0 | 0 |
| F2 | 6.97e-29 | 7.49e-28 | 7.96e-29 | 0.01425 | 3.58e-15 | 0.0455 | **0** | 3.24e-11 |
|  | 1 | 1 | 1 | 1 | 1 | 1 | **1** | 1 |
|  | 0 | 0 | 0 | 0 | 0 | 0 | **0** | 0 |
| F3 | 885090 | 725750 | 946043 | 6.49e+06 | **1.96e-15** | 5.06e+06 | 2.65e+07 | 90.6979 |
|  | 1 | 1 | 1 | 0.48 | **1** | 0.44 | 1 | 1 |
|  | 0 | 0 | 0 | 7.87e-05 | **0** | 6.51e-05 | 0 | 0 |
| F4 | **13.5728** | 13.7135 | **13.5728** | 40.796 | 2.587 | **13.5728** | 13.9880 | 158.418 |
|  | **1** | 1 | **1** | 1 | 1 | **1** | 1 | 1 |
|  | **0** | 0 | **0** | 0 | 0.7517 | **0** | 0 | 0 |
| F5 | 8.35e-29 | 0.3189 | 8.92e-29 | 10.9402 | 21.2546 | 16.6896 | **0** | 14.9547 |
|  | 1 | 1 | 1 | 1 | 1 | 1 | **1** | 1 |
|  | 0 | 0 | 0 | 0 | 0 | 0 | **0** | 0 |
| F6 | 7941.59 | 7154.11 | 7946.24 | 7615.43 | 4.7360 | 7550.13 | 8600.818 | **687.2246** |
|  | 0.4 | 0 | 0 | 0 | 0.2 | 0 | 0 | **0.96** |
|  | 0 | 0.0116 | 0.01264 | 0.2675 | 0.9332 | 0.3818 | 0.01525 | **5.85e-06** |
| F7 | -145.62 | **-260.388** | -172.159 | -21.6109 | -2563.96 | -13.0956 | -153.963 | -941.8285 |
|  | 1 | **1** | 1 | 0.04 | 0 | 0 | 1 | 0.88 |
|  | 0 | **0** | 0 | 0.00102 | 5011.44 | 0.00155 | 0 | 1.66e-05 |
| F8 | **-0.00013** | **-0.00013** | **-0.00013** | 0.00366 | -14.9387 | 0.04298 | 0.000286 | 0.0002 |
|  | **1** | **1** | **1** | 0.96 | 0 | 1 | 1 | 1 |
|  | **0** | **0** | **0** | 4.76e-06 | 1.82e+07 | 0.00298 | 0 | 0 |
| F9 | **-0.00204** | **-0.00204** | **-0.00204** | -0.00203 | 0.7054 | **-0.00204** | -0.00168 | **-0.00204** |
|  | **1** | **1** | **1** | 1 | 0 | **1** | 1 | **1** |
|  | **0** | **0** | **0** | 0 | 21419.9 | **0** | 0 | **0** |
| F10 | **-4.83e-05** | **-4.83e-05** | **-4.83e-05** | 5.07e-05 | -16.7263 | -3.24e-05 | 9.12e-05 | 5.81e-05 |
|  | **1** | **1** | **1** | 1 | 0 | 1 | 1 | 1 |
|  | **0** | **0** | **0** | 0 | 8.33e+07 | 0 | 0 | 0 |
| F11 | -1.4199 | 1.7584 | -1.6628 | 48.59 | -4946.35 | 65.2575 | **-1.9254** | -160.1565 |
|  | 0.92 | 0.44 | 1 | 0 | 0 | 0 | **1** | 0 |
|  | 1.83e-32 | 6.11e-12 | 0 | 2.26e+50 | 2.70e+99 | 2.4313 | **0** | 0.3887 |
| F12 | 51.6255 | 27.5311 | 53.3338 | **4.0002** | 11.3024 | 4.1081 | 7.3566 | 12.9012 |
|  | 1 | 1 | 1 | **1** | 0 | 1 | 1 | 1 |
|  | 0 | 0 | 0 | **0** | 4.5132 | 0 | 0 | 0 |
| F13 | **23.2901** | 51.9073 | 26.824 | 126.221 | 10.5261 | 70.7797 | 73.6353 | 1293.788 |
|  | **1** | 1 | 1 | 1 | 0.8 | 1 | 1 | 1 |
|  | **0** | 0 | 0 | 0 | 1.58983 | 0 | 0 | 0 |
| F14 | 1.4330 | 1.5756 | 1.4119 | 1.4327 | 0.3148 | 1.6326 | 1.4611 | **1.1840** |
|  | 1 | 1 | 1 | 1 | 0 | 1 | 1 | **1** |
|  | 0 | 0 | 0 | 0 | 2.5679 | 0 | 0 | **0** |
| F15 | 18.9438 | 20.4518 | 17.5615 | 20.7031 | 31.4785 | 20.3261 | 13.6659 | **11.2783** |
|  | 0.96 | 0.12 | 1 | 1 | 0 | 1 | 1 | **1** |
|  | 2.63e-06 | 0.00016 | 0 | 0 | 2279.59 | 0 | 0 | **0** |
| F16 | 256.542 | 281.738 | 260.124 | 375.734 | **4.76e-15** | 302.85 | 251.4531 | 13.069 |
|  | 1 | 1 | 1 | 0.92 | **1** | 1 | 1 | 1 |
|  | 0 | 0 | 0 | 4.78e-06 | **0** | 0 | 0 | 0 |
| F17 | 1.0384 | 1.0416 | **1.0312** | 1.0468 | 0.00856 | 1.0492 | 1.0450 | 1.0480 |
|  | 0 | 0 | **0** | 0 | 0 | 0 | 0 | 0 |
|  | 25.5 | 25.5 | **25.5** | 25.5 | 118.433 | 25.5 | 25.5 | 25.5 |
| F18 | 854.438 | 9581.65 | 6219.43 | 857.646 | **6.6** | 14321.2 | 7340.76 | 20155.975 |
|  | 0 | 0 | 0 | 0 | **0** | 0 | 0 | 0 |
|  | 134670 | 1.51e+07 | 4.05e+07 | 18625.7 | **5542.67** | 5.46e+07 | 8.96e+06 | 6.56e+08 |
| F19 | 1.21e-05 | 1.51e-05 | 1.21e-05 | 6.52e-06 | -46.041 | 1.49e-05 | **0** | 8.8444 |
|  | 0 | 0 | 0 | 0 | 0 | 0 | **0** | 0 |
|  | 36116.2 | 36116.2 | 36116.2 | 36116.2 | 36185.7 | 36116.2 | **36116.2** | 36130.22 |
| F20 | 3.1344 | **2.5485** | 3.2117 | 6.2603 | 3.2270 | 7.7770 | 4.5789 | 6.8396 |
|  | 1 | **1** | 1 | 1 | 0.96 | 1 | 1 | 1 |
|  | 0 | **0** | 0 | 0 | 0.004264 | 0 | 0 | 0 |
| F21 | 62.1813 | 12.2582 | 63.3177 | 25.4192 | 235.007 | 57.1599 | 62.1685 | **6.8466** |
|  | 1 | 1 | 1 | 1 | 0 | 1 | 1 | **1** |
|  | 0 | 0 | 0 | 0 | 114.908 | 0 | 0 | **0** |
| F22 | **2997.47** | 64028.9 | 8051.21 | 74216.5 | 85.3516 | 40693 | 7838.53 | 3186.796 |
|  | **1** | 0.28 | 0.96 | 0 | 0.76 | 0 | 1 | 0.76 |
|  | **0** | 3.0501 | 0.0056 | 51.4983 | 217.017 | 43.6401 | 0 | 0.2074 |
| F23 | 1.3788 | 1.5710 | 1.3257 | 1.5441 | 21.2107 | 1.5225 | 1.3742 | **1.1240** |
|  | 1 | 1 | 1 | 1 | 0 | 1 | 1 | **1** |
|  | 0 | 0 | 0 | 0 | 443239 | 0 | 0 | **0** |
| F24 | 14.1685 | 17.0588 | 13.9172 | 75.9954 | 10.2124 | 17.9385 | 14.9225 | **11.6550** |
|  | 1 | 1 | 1 | 0 | 0 | 1 | 1 | **1** |
|  | 0 | 0 | 0 | 21883.1 | 277.846 | 0 | 0 | **0** |
| F25 | 252.584 | 299.896 | 253.526 | 609.72 | 51.0171 | 315.039 | 254.1548 | **27.52** |
|  | 1 | 1 | 1 | 0 | 0 | 1 | 1 | **1** |
|  | 0 | 0 | 0 | 3346.36 | 0.8301 | 0 | 0 | **0** |
| F26 | 1.0389 | 1.0435 | **1.0366** | 1.0475 | 0.003844 | 1.04887 | 1.0411 | 1.0478 |
|  | 0 | 0 | **0** | 0 | 0 | 0 | 0 | 0 |
|  | 25.5 | 25.5 | **25.5** | 25.5 | 117.822 | 25.5 | 25.5 | 25.5 |
| F27 | 4866.84 | 19907 | 13404.6 | **818.044** | 416.528 | 48022.1 | 6.37e+04 | 60198.128 |
|  | 0 | 0 | 0 | **0** | 0 | 0 | 0 | 0 |
|  | 1.53e+07 | 2.33e+08 | 2.00e+07 | **69432** | 149219 | 1.84e+09 | 2.00e+08 | 6.52e+09 |
| F28 | 213.915 | 268.093 | 269.195 | 275.074 | 91.0368 | 248.88 | 267.1889 | **129.11** |
|  | 0 | 0 | 0 | 0 | 0 | 0 | 0 | **0** |
|  | 36273 | 36324 | 36316.3 | 36328 | 36294.1 | 36321.5 | 36317.4 | **36258.2** |
| +/=/− Criterion I | 20/3/5 | 14/5/9 | 21/0/7 | 24/0/4 | 23/2/3 | 18/0/10 | 16/1/11 |  |
| +/=/− Criterion II | 18/7/3 | 8/20/0 | 22/2/4 | 21/2/5 | 23/4/1 | NA | NA |  |

Optimization (SajDE) [39].

7) The standard differential evolution algorithm (DEbin) [6].

Two evaluation criteria are used for performance comparison for more accurate evaluation. The first evaluation criterion is referenced in the guidelines of CEC2017 [32] and employed in results of all comparison algorithms. The detailed description of the evaluation criterion is showed at the beginning of Section IV. The sign test [38] is also used to judge the significance of the results (CAL-SHADE, L-SHADE44, SaDE, SajDE and DEbin versus APDE-NS and APDE-NS-L) at the 0.05 significance. The symbols "+", "−", "=" are employed to indicate APDE-NS perform significantly better ( + ), no significantly difference (=), significantly worse (−) than the algorithms in comparison. Tables III-V show the comparison results using the two evaluation criteria. The comparison results using the two evaluation criteria in 10-D problems are presented in Table III in terms of SR, $\overline{vio}$ and mean, the comparison in 30-D problems is listed in Table IV, the comparison results in 50-D problems are showed in Table V. For clarity, the best evaluation indicator is highlighted in **boldface**. The data in these tables show the mean, FR and $\overline{vio}$ from top to bottom. The results of L-SHADE44+IDE and UDE are from [34] [36]. Moreover, in order to make the algorithm comparison results intuitive and fairly compared with other algorithms, successful optimization rate (SR) of each algorithm for the entire benchmark functions set is listed in Table I. When an algorithm can obtain a reliable feasibility rate (FR>0.3) in one benchmark function, we consider that the benchmark function is optimized successfully by the algorithms. From Table I, it can be seen that APDE-NSs perform better than other algorithms and it can successfully search for the feasible solution with a high reliability. APDE-NSs only fail on $f_{17}, f_{18}, f_{19}, f_{26}, f_{27}$ and $f_{28}$. All mentioned algorithms in the paper also fail to search for the feasible solution to the above seven benchmark functions. For $f_6$, only APDE-NSs and UDE can find feasible solution steadily. UDE is invalid when optimizing $f_{11}$, however, APDE-NSs can solve the function efficiently.

TABLE VI. COMPARISON RESULTS OF APDE-NS AND ITS VARIANTS WITH SINGLE MUTATION STRATEGY ON 30-D PROBLEMS

| Func/30D | APDE-NS-A | APDE-NS-B |
|---|---|---|
| F1 | + | + |
| F2 | + | + |
| F3 | = | + |
| F4 | = | = |
| F5 | = | + |
| F6 | = | + |
| F7 | = | + |
| F8 | = | + |
| F9 | = | = |
| F10 | = | + |
| F11 | = | = |
| F12 | + | = |
| F13 | = | = |
| F14 | − | + |
| F15 | = | = |
| F16 | = | + |
| F17 | − | + |
| F18 | + | + |
| F19 | + | + |
| F20 | = | + |
| F21 | = | − |
| F22 | + | + |
| F23 | = | − |
| F24 | = | + |
| F25 | = | + |
| F26 | − | + |
| F27 | + | + |
| F28 | + | + |
| + | 8 | 20 |
| = | 17 | 6 |
| − | 3 | 2 |

From Table III-V, the performance of our proposed APDE-NSs is superior to or at least competitive with other algorithms:

*1) Under the first evaluation criterion*

For the comparison result with 10-D, as demonstrated in Table III, APDE-NS-L is superior to or at least as good as the other algorithms. APDE-NS-L performs better than the ranked first algorithm of the CEC2017 constrained optimization (L-SHADE-44). APDE-NS-L performs better than L-SHADE-44 on 15 functions, while beaten by L-SHADE-44 on 7 functions. Moreover, our proposed APDE-NS-L is efficient at solving $f_6$, while L-SHADE-44 performs poor on the function. On 30-D problems, as shown in Table IV, APDE-NS performs better than or at least equal to the other algorithms and keep the ability to solve $f_6$. APDE-NS-L is superior to L-SHADE-44 in 13 functions. For 50-D problems, as presented in Table V, APDE-NS outperforms the other algorithms in most benchmark functions. APDE-NS get competitive results that are equal or better than L-SHADE-44 in 28 benchmark functions. In summary, APDE-NSs demonstrate comprehensiveness in constrained optimization superior to the other algorithms. Except $f_{17}, f_{18}, f_{19}, f_{26}, f_{27}, f_{28}$ (all mentioned algorithm in the paper is invalid for these functions), APDE-NS can optimize all other functions. The other algorithms are inefficient for it. In addition, in the comparison algorithm, UDE is also effective in optimizing $f_6$, however, it fails to optimize $f_{11}$. As a result, our proposed algorithm has an excellent constraint control capability and can follow the suitable balance point between exploration and exploitation during evolutionary process through state switching mechanism and population size adaptive mechanism.

*2) Under the second evaluation criterion*

When we employ the sign test method at $\alpha=0.05$ to evaluate the statistical significance of the comparison results between APDE-NSs and other state-of-the-art algorithms, if APDE-NSs can win more than 18 independent runs in all 25 independent runs with respect to one benchmark function, we consider APDE-NSs perform significantly better than the corresponding comparison algorithm. From the Table III-V, as we can see, for 10-D problems, APDE-NS-L performs better than or at least equal to the other algorithms. For the comparison result with 30-D, the result of APDE-NS is statistically superior to L-SHADE-44 in 4 functions and it is beaten in 3 functions, while APDE-NS is statistically equal to or better than L-SHADE-44 on all 50-D benchmark functions. APDE-NS outperforms in 8 functions and equal in 20 functions, APDE-NS is not beaten by L-SHADE-44 in any function. The increase in problem dimension does not affect the performance of APDE-NS, even the performance of APDE-NS is better than that of 30-D problems compared to the other comparison algorithms.

In general, we can conclude that the performance of APDE-NSs is superior to or at least competitive with other state-of-the-art algorithms in comparison under both evaluation criteria. Besides, the experimental result shows that APDE-NSs can follow the changing suitable balance point between exploration and exploitation and have stronger constraint control capabilities.

*D. Effects of APDE-NS Mechanisms*

In this part, two core mechanisms (combination mechanism of State-switch DE/current-to-pbest/1 strategy and State-switch

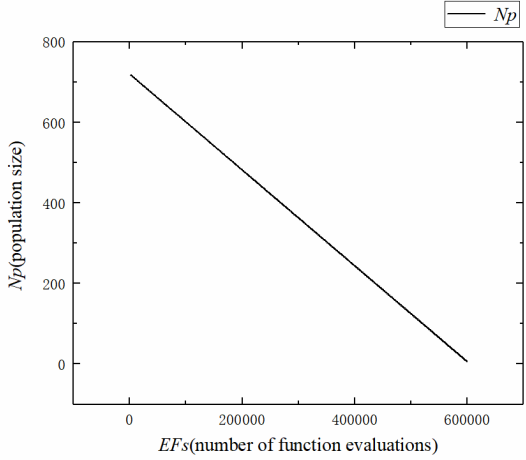

Fig. 6. Population size adaptive process when solving low strict level constraints $f_1$, 30-D

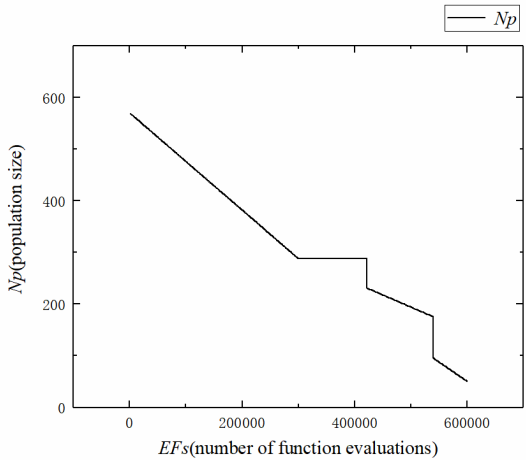

Fig. 7. Population size adaptive process when solving medium strict level constraints $f_7$, 30-D

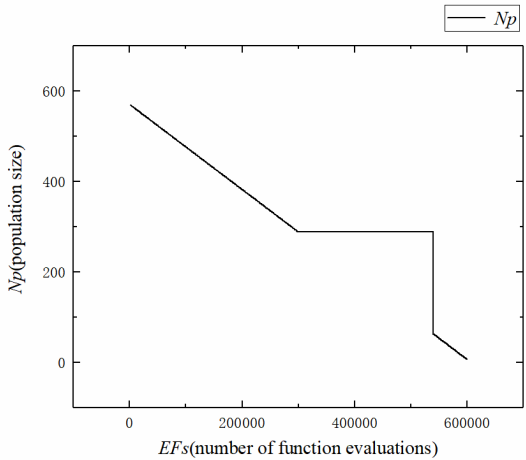

Fig. 8. Population size adaptive process when solving high strict level constraints $f_{17}$, 30-D

DE/randr1/1 strategy and adaptive population size mechanism based on *pfeas*) of our proposed APDE-NS are segmented and run separately. For space limitations, we chose 30-D problems to analyze the effects of each mechanisms. The detailed analysis of the experiment are as follows:

*1)* The analysis of the former mechanism

We consider two APDE-NS variants with single DE/current-to-pbest/1 strategy or DE/randr1/1 strategy and call algorithms modified in the above manner as APDE-NS-A and APDE-NS-B. APDE-NS-A only use the DE/current-to-pbest/1 strategy and APDE-NS-B only use DE/randr1/1 strategy. The rest components of these two are consistent with APDE-NS. The evaluation criteria employ the first method. The comparison results are reported in Table VI. The symbol "+", " − ", "=", respectively, denotes the performance of APDE-NS statistically superior to, equal to, worse than the compared algorithm.

From comparison results listed in Table VI, we can see our proposed combination mechanism of State-switch DE/current-to-pbest/1 strategy and State-switch DE/randr1/1 strategy improves the algorithm performance compared to the ordinary single strategy without state switching technique significantly. It affirms the effectiveness of the proposed mechanism.

*2)* The analysis of the latter mechanism

APDE-NS employs adaptive population size mechanism is based on feasible rate. In order to discuss the effect of this mechanism on the performance of the algorithm. First, we analyze population size adaptive process when dealing with different strict level constraints. Details as shown in Fig. 6-8.

As we can see from Fig. 6, population size keep linear reduction in optimization process when solving low strict level constraints. Population size reduction linearly is suitable for solving low strict level constraints like $f_1$. The feasible rate is large enough in the early stage of evolution and population size reduction linearly can provide high exploitation in the middle and late stages of evolution to avoid premature and speed up convergence. When solving medium strict level constraints like $f_7$, as Fig. 7 depicts, the algorithm fails to find a feasible solution during the exploring biased state, so population size stops decreasing and constant. It provides high diversity to search for feasible solution. When first feasible solution is found, population size continues decreasing to get higher exploitation and the rate of decline is based on *pfeas*. As Fig. 8 shows, for high strict level constraints, our algorithm fails to find feasible solution until end of the balance search state. Algorithm reduces population size directly to a low-level to get high exploration for better objective function value.

In this part, in order to evaluate the effectiveness of the novel mechanism, the APDE-NS variants with constant population size are compared with APDE-NS on 30-D problems. The results are presented in Table VII in detail. The meaning of symbol "+", " − ", "=" is the same as the former mechanism. From the results listed in Table VII, we can see adaptive population size mechanism based on feasible rate improves the performance of the algorithm significantly.

## V. CONCLUSION

This paper proposes an adaptive population size differential evolution with novel mutation strategy for constrained optimization problem. The proposed algorithm, namely, APDE-NSs achieve adaptive adjustment of the balance between exploration and exploitation. In order to follow the change of suitable balance point between exploration and exploitation, two novel mechanism are proposed in the paper: 1) a set of state-switch mutation strategies; 2) adaptive population size based on *feas* (feasible rate). The state-switching mechanism can strengthen constraint control ability and improve the adaptability to solve different types of problems.

TABLE VII. COMPARISON RESULTS OF APDE-NS AND ITS VARIANTS WITH CONSTANT POPULATION SIZE

| Func/30D | APDE-NS-5NP | APDE-NS-10NP | APDE-NS-15NP | APDE-NS-20NP |
|---|---|---|---|---|
| F1  | = | = | = | + |
| F2  | + | = | = | + |
| F3  | = | = | = | = |
| F4  | + | = | + | + |
| F5  | = | = | + | + |
| F6  | + | + | + | = |
| F7  | = | = | = | = |
| F8  | = | = | = | = |
| F9  | = | = | = | = |
| F10 | = | = | = | = |
| F11 | = | = | = | = |
| F12 | + | = | − | = |
| F13 | = | + | + | + |
| F14 | = | = | = | = |
| F15 | + | = | = | = |
| F16 | = | = | = | = |
| F17 | = | = | = | = |
| F18 | = | + | + | + |
| F19 | + | + | = | − |
| F20 | + | + | + | + |
| F21 | = | = | = | = |
| F22 | + | + | = | + |
| F23 | = | = | + | + |
| F24 | = | = | = | = |
| F25 | = | = | = | = |
| F26 | = | = | = | = |
| F27 | = | + | + | + |
| F28 | + | + | = | = |
| +   | 9 | 8 | 8 | 10 |
| =   | 19 | 20 | 19 | 17 |
| −   | 0 | 0 | 1 | 1 |

From the results of comparison with other state-of-the-art algorithms, APDE-NSs show a better or at least competitive performance compared to other algorithms. The experimental results also indicate that the state-switching mechanism can improve the performance of the algorithm. However, APDE-NSs can not solve $f_{17}, f_{18}, f_{19}, f_{26}, f_{27}, f_{28}$, but other algorithms in this paper also fail to search feasible solutions. Moreover, the number of benchmark functions APDE-NSs can find feasible solution is most compared to other algorithms.

In this paper, our main aim is to improve the algorithm's control over constraints. Therefore, the optimization of the objective function value is the weakness of our algorithm. For future work, we will extend the APDE-NSs to keep a suitable balance between constraint control and objective function values. It can improve the performance of the algorithm in solving real-world applications.